\pdfoutput=1

\documentclass[11pt]{article}
\usepackage[dvipsnames]{xcolor}

\usepackage{EMNLP2023}
\usepackage[english]{babel}

\usepackage{times}
\usepackage{latexsym}
\usepackage{arydshln}

\usepackage{CJKutf8}

\usepackage{rotating}

\usepackage[OT2,T1]{fontenc}

\newcommand\textcyr[1]{{\fontencoding{OT2}\selectfont #1}}

\usepackage{tipa}

\usepackage[utf8]{inputenc}

\usepackage{microtype}

\usepackage{booktabs}

\usepackage{graphicx}
\usepackage{float,lscape}
\usepackage{array}
\usepackage{xspace}
\usepackage{inconsolata}

\newcommand{\langcount}{55}
\newcommand{\pipeline}{\mbox{\textsc{Gender-GAP}}\xspace}

\newcommand{\nllb}{NLLB}

\newcommand\blfootnote[1]{%
  \begingroup
  \renewcommand\thefootnote{}\footnote{#1}%
  \addtocounter{footnote}{-1}%
  \endgroup
}

\title{The \pipeline Pipeline: A Gender-Aware Polyglot Pipeline\\ for Gender Characterisation in \langcount{} Languages}

\author{Benjamin Muller, Belen Alastruey, Prangthip Hansanti, Elahe Kalbassi, \\ \textbf{Christophe Ropers, Eric Michael Smith, Adina Williams, Luke Zettlemoyer,} \\ \textbf{Pierre Andrews$^*$ and Marta R. Costa-jussà$^*$} \\
Meta AI \\
  \texttt{\{benjaminmuller,alastruey,prangthiphansanti,ekalbassi,chrisopers} \\
    \texttt{ems,adinawilliams,lsz,mortimer,costajussa\}@meta.com} \\}

\begin{document}
\maketitle
\begin{abstract}
\blfootnote{$^*$ Equal Research Leadership Contribution}
Gender biases in language generation systems are challenging to mitigate. 
One possible source for these biases is gender representation disparities in the training and evaluation data. 
Despite recent progress in documenting this problem and many attempts at mitigating it, we still lack shared methodology and tooling to report gender representation in large datasets. Such quantitative reporting will enable further mitigation, e.g., via data augmentation. This paper describes the \pipeline Pipeline (for \textbf{G}ender-\textbf{A}ware \textbf{P}olyglot Pipeline), an automatic pipeline to characterize gender representation in large-scale datasets for \langcount{} languages. The pipeline uses a multilingual lexicon of gendered person-nouns to quantify the gender representation in text. We showcase it to report gender representation in  WMT\footnote{http://www2.statmt.org/wmt23/} training data and development data for the News task, confirming that current data is skewed towards masculine representation. Having unbalanced datasets may indirectly optimize our systems towards outperforming one gender over the others. We suggest introducing our gender quantification pipeline in current datasets and, ideally, modifying them toward a balanced representation.\footnote{The \pipeline pipeline is available at \url{https://github.com/facebookresearch/ResponsibleNLP/tree/main/gender_gap_pipeline}}
\end{abstract}

\section{Introduction}
\label{sec:intro}

Despite their widespread adoption, Natural Language Processing (NLP) systems are typically trained on data with social and demographic biases. Such biases inevitably propagate to our models and their generated outputs, e.g., by over-representing some demographic groups and under-representing others. It is, therefore, critical to measure, report, and design methods to mitigate these biases, before they can be encoded or even amplified during training \citep{intersectional_fairness,wang2021directional}.

\begin{figure}[t!]
    \centering
    \includegraphics[width=0.35\textwidth]{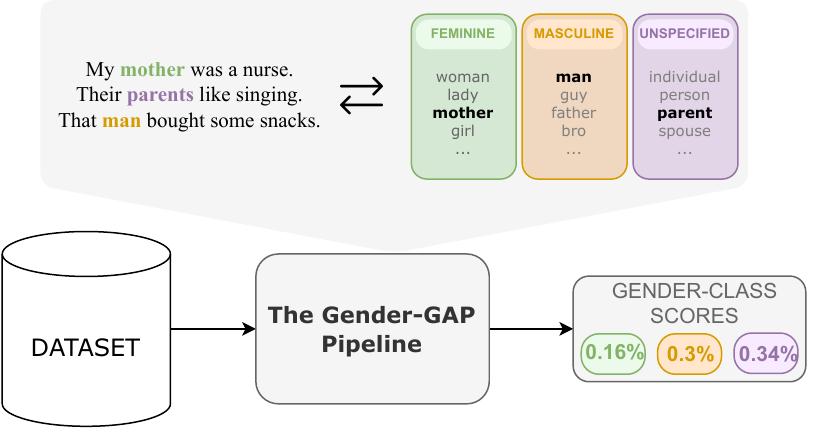}
    \caption{The Gender-GAP Pipeline works by identifying gendered lexical terms and reporting statistics on these lexical matching.}
    \label{fig:pipeline}
\end{figure}

This paper focuses on quantifying gender representation in highly multilingual data (see Figure \ref{fig:pipeline}), in particular, for the task of machine translation. %
Gender is a complex concept that can be defined in many ways depending on the field of study, language or culture \citep{Chandra:81,hellinger2001gender,close_look_gender_assignement}. We discuss and define gender in Section~\ref{sec:genders}. 
However, briefly, we define gender bias as the systematic unequal treatment based on one’s gender \citep{blodgett-etal-2020-language,stanczak2021survey}. %
Gender bias, when it impacts training data, may decrease the performance of the system on certain gender groups \citep{hovy-etal-2020-sound}. When impacting evaluation data, it may push the system designers to deploy a system that causes harm by favoring one group over others \citep{survey_acm_2021}. 
For example, a system that translates text that includes feminine nouns more poorly than text with masculine nouns may lead the end users to miss important information or misunderstand the sentence \citep{savoldi-etal-2021-gender}. A system that inaccurately translates a gender-neutral sentence in English e.g. \textit{they are professors} to a sentence with a masculine noun \textit{ils sont professeurs} in French may also lead to serious representational harm.  %

We propose the \pipeline pipeline to quantify gender representation bias of multilingual texts using lexical matching as a proxy.  
Our pipeline can be seen as two main modules. 

First, we build a multilingual gender lexicon: starting from a list of about 30 English nouns extracted from the HolisticBias dataset \citep{smith-etal-2022-im}, split into 3 gendered classes---masculine, feminine, and unspecified. We manually translate them and reassign them to the appropriate gender class for each target language (e.g. ``grandfathers'', masculine in English, becomes ``abuelos'', masculine and unspecified in Spanish). Our list is restricted to nouns that refer to people (e.g. man, woman, individual) or to kinship relationships (e.g. dad, mom, parent). Most languages, including genderless languages \citep{PrewittFreilino2012TheGO} (e.g. Finnish, Turkish) encode genders through kinship relationships and person terms \citep{savoldi-etal-2021-gender}. For this reason, focusing on a restricted list of kinship and person nouns allow us to scale our lexicon to \langcount{} languages.  %

Second, we arrive at a straightforward and easily comparable gender distribution by using a word matching counter. Based on our newly collected multilingual lexicon, our pipeline segments each input sentence at the word-level using Stanza \citep{qi-etal-2020-stanza}, a state-of-the-art word segmentation tool, and counts the number of occurrences of words in each gender class.  As a result, we obtain a gender distribution across \langcount{} languages. %

In summary, our contribution is threefold: 
\begin{itemize}
    
    \item We build an aligned multilingual lexicon that can support measurement of the representation of genders in \langcount{} languages.
    \item We introduce the Gender-Aware Polyglot pipeline (\pipeline), a lexical matching pipeline, and describe the gender distribution observed in popular machine translation training and evaluation data. On average, all three analyzed datasets are biased toward the masculine gender. We find the gender representations to be domain- and language-specific. Additionally, using the \pipeline pipeline, we can discover sentences that have been translated with a gender bias.
    \item We release our pipeline and recommend the reporting of gender representations in machine translation training and evaluation datasets to improve awareness on potential gender biases.

\end{itemize}

\section{Related work}

The study of biases in text has become more important in recent years, with Large Language Models (LLMs)  displaying bias against people depending on their demographics and identity. 
As a testament to the importance of this topic, many recent papers, including those introducing GPT-3 and 4 \citep{brown2020language,OpenAI2023GPT4TR}, PaLM 1 and 2 \citep{chowdhery2022palm,anil2023palm}, LLaMa 1 and 2 \citep{touvron2023llama,touvron2023llama2}, %
analyze how such biases affect their model outputs. Some works even discuss frequencies of gendered terms in their pretraining corpora \citep{anil2023palm,touvron2023llama2}, as this can affect downstream generation. %
Despite this acknowledgment of the issue, general purpose tools to measure demographic biases are still fairly rare, and so far have mainly been in English. 

However, some have begun to measure demographic biases beyond English. \citet{smith-etal-2022-im} built a comprehensive analysis dataset covering 13 demographic groups and \citet{costa2023multilingual} extended it to the multilingual setting. Specific to Machine Translation, \citet{savoldi-etal-2021-gender} discussed best practices in reporting gender bias. Several works \citep{stanovsky-etal-2019-evaluating, prates-etal-2020-assessing, renduchintala-etal-2021-gender, renduchintala-williams-2022-investigating} have explored metrics for exposing failures in automatically translating pronoun and occupations, and some have even explored MT model training \citep{escude-font-costa-jussa-2019-equalizing,stafanovics-etal-2020-mitigating} or fine-tuning \citep{saunders-etal-2020-neural, corral-saralegi-2022-gender,costa-jussa-de-jorge-2020-fine} or both \citep{choubey-etal-2021-gfst} to lessen the effect of gender-related biases. More than this, there are initiatives that provide toolkits to generate multilingual balanced datasets in terms of gender \cite{Costajuss2019GeBioToolkitAE} from Wikipedia and even balanced in gender within occupations \cite{NEURIPS2022_09933f07}. %

However, despite the progress made, most of these resources only cover a handful of languages---the community still lacks easy to use, open-source toolkits to measure biases across a large number of languages. 
In this work, we address this need by %
showcasing, \pipeline, a lexical matching pipeline to measure gender distribution across \langcount{} languages.

\begin{figure}[!t]
    \centering
    \includegraphics[width=0.48\textwidth]{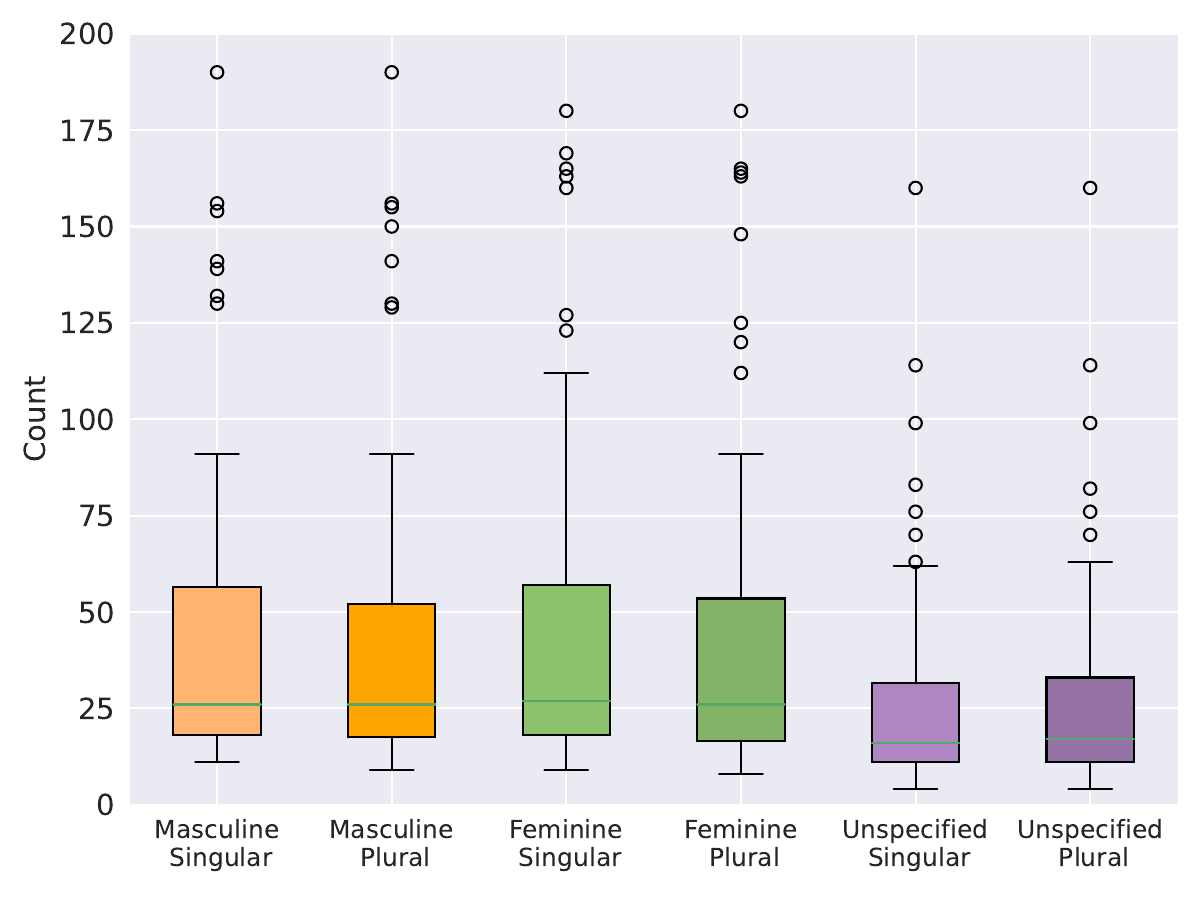} 
    \caption{Distribution of the words in our proposed dataset across different languages, gender-classes and number.}
    \label{fig:languages_barplot}
\end{figure}

\section{Proposed Data Collection and Pipeline}
\label{sec:datacollection}

\subsection{Defining Gender}
\label{sec:genders}

Gender is a complex topic that can be defined in many different ways depending on the field of studies and the context \citep{hellinger2001gender}. We approach gender from two perspectives: %

First, linguistic gender \cite{wals-30,cao-daume-iii-2020-toward,close_look_gender_assignement,stanczak2021survey} corresponds to the classification of linguistic units, such as words, into categories based on the gender information they provide. Linguistic gender refers to overlapping notions, such as \textit{grammatical}, and \textit{semantic} gender, depending on the properties of the language. Grammatical gender implies the classification of nouns, adjectives, and other parts of speech into categories based on their morphosyntactic properties (e.g., in Russian masculine nouns typically end in \textcyr{-й}, feminine nouns typically ends in -a or \textcyr{–я} and neuter nouns usually end in \textcyr{-о}, \textcyr{-е}, or \textcyr{-ё}). In many languages, grammatical gender morphology appears on all nouns, regardless of whether they refer to persons, animals, plants, or inanimate objects (e.g., ``il libro'' \textit{the book} is a masculine noun in Italian). Semantic gender \citep{corbett_1991} 
refers to the existence of lexical units whose meaning is associated with a specific cultural notion of peoples' gender(s). For instance, in English, the word ``men'' associated with masculine traits, ``woman'' with feminine ones, etc. Semantic gender then may be present in languages that do not morphologically mark grammatical gender, such as English, Turkish, or Mandarin Chinese. In languages that do mark grammatical gender, grammatical and semantic gender do not always match: for example, in German, the word for girl ``M\"{a}dchen'' is grammatically neuter, but refers to a person which would fall into our 'feminine' class based on its meaning. For our purposes, we use semantic gender classes in our multilingual lexicon, since we are interested in gender representation.

Our goal is to build and foster inclusive NLP technologies that do not carry, replicate, or amplify social gender biases, which can impact end users and societies negatively by affecting representations of specific groups. However, there are social meanings of gender that are not readily accessible in text, so, we use semantic gender on human words as a proxy for social gender. 

Social gender refers to gender as a social construct based on cultural norms and identity \citep{ackerman2019syntactic,cao-daume-iii-2020-toward,stanczak2021survey,duignan2023gender}. As highlighted in \citeauthor{ackerman2019syntactic}, social gender is defined as the internal gender experienced by a given human individual. For this reason, data-driven analysis of genders in large corpora can only relate to social gender indirectly through linguistic notions of gender(s).\footnote{We recall that gender is distinct from sex which refers to collections of biological properties of individuals such as genes (e.g., chromosomes), phenotypes (e.g., anatomy) \cite{COE2023}. See \citet{butler-2011-bodies} for a discussion of additional factors that complicate this view.}
We assume for our purposes that a list of gendered words can be used to approximate some important aspects of social gender for the purposes of measuring representation disparities.

\subsection{Aligned Gendered Multilingual Lexicon}
\label{sec:lexicon}

To measure gender distribution across \langcount{} languages, we first build a multilingual lexicon.  We want this lexicon to be as aligned as possible across languages while also encoding language-specific gender linguistic phenomena. %

\begin{figure*}[t!]
    \centering
    \includegraphics[width=\textwidth]{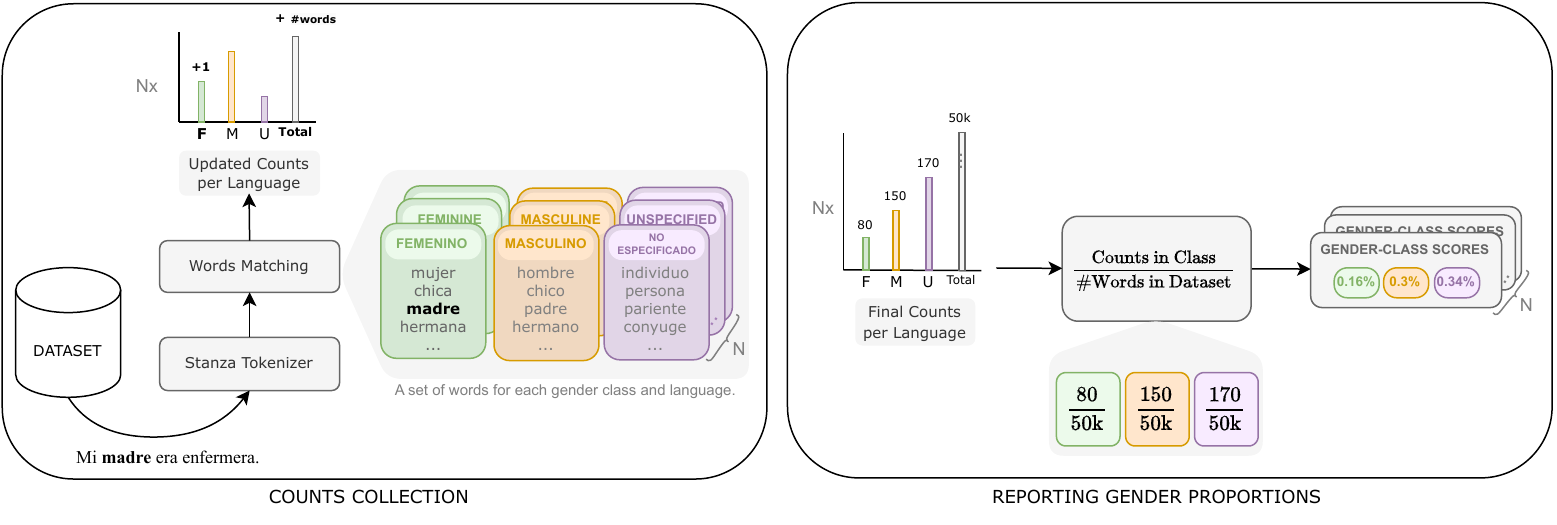} 
    \caption{Diagram of the \pipeline pipeline. In the first stage, we process each sentence of the \langcount~supported languages of the dataset and count the word matches for each category. Once this step is completed, we compute a gender-class score which corresponds to the proportion of gendered noun matched within all the words in the dataset. %
    }
    \label{fig:pipeline_diagram}
\end{figure*}

\paragraph{Languages} Our lexicon is available in 55 typologically and phylogenetically diverse languages such as English, Finnish, Zulu, Vietnamese, Ganda, Japanese or Lithuanian, spanning 15 distinct scripts. We report the complete list of languages in Figure~\ref{table:language_list}.

\paragraph{Gender Classes} We define three semantic gender classes: masculine, feminine and unspecified. %
The unspecified class aggregates nouns of different sorts. It mainly capture nouns that do not explicitly encode any particular gender (e.g. ``person'' is considered unspecified in English). For this reason, ``unspecified'' can be seen as aggregating masculine, feminine and non-binary genders \citep{herdt2020third}. 

While there exist more complex gender lexica as discussed in~\citet{stanczak2021survey}, they are focused on English and are not always easily translated. Because our goal is to provide a methodology that can be used to evaluate bias across multiple languages, we take a more pared down lexical approach.

\paragraph{Lexicon creation} We start by defining a list of about ten, high frequency person nouns per gender class in English. Each noun is found in both its singular and plural form. To find a list of nouns that is as universal as possible, we restrict this list of persons such as masculine ``man'', feminine ``woman'', and ``person'' and synonyms (e.g. ``individual'') that we complement with kinship terms classified by gender (e.g., masculine ``father'', feminine ``mother'', neutral ``parent''). %
Our list corresponds to the one defined in the previous work of HolisticBias \citep{smith-etal-2022-im}, which is only available in English.\footnote{We use the gender noun list v1.1 from HolisticBias}%

We then translate these nouns into the other languages by reassigning them to the appropriate gender class. %
A noun in a given gender class may be part of another class (or multiple other classes) in another language. For instance ``grandparents'' (masculine, plural) becomes ``abuelos'' in Spanish which is both masculine and unspecified genders.

The English-language source list is passed on translators who are native speakers of the target language, with language proficiency at CEFR\footnote{\url{https://coe.int/en/web/common-european-framework-reference-languages/level-descriptions} retrieved 2023-07-24} level C2 in the source language. For all languages, translators are asked to provide equivalent singular and plural terms in their respective native language, except if any of the source concepts do not exist in the language. For example, not all languages use a distinctive, gender-agnostic term such as the English term \textit{sibling}, distinctively from either \textit{brother} or \textit{sister}.
 We also consider that the reverse can be true (i.e. that the target language may have more than one term to translate one of the English terms in the source list), and give the translators the possibility to provide additional translations in such cases. For instance, when we translate \textit{women} into Korean we get :\begin{CJK}{UTF8}{} \CJKfamily{mj} ``여성들'' \end{CJK} and \begin{CJK}{UTF8}{} \CJKfamily{mj} ``여인들''\end{CJK}. 
 
 Additionally, translators are asked to consider the terms in the source list as lemmas (or headwords in dictionary entries) and, if applicable to the given language, to provide relevant morphologically derived forms, including cases and gendered forms. Finally, translators are also encouraged to provide terms covering all language registers, which is necessary because some languages (e.g., Thai or Korean, among others) use several different terms at various levels of formality.

We are cognizant of the fact that this approach presents several limitations. The first limitation occurs when a term could be said to fall into both the unspecified and one of the gendered categories. For example, the term Spanish \textit{padres} can be used to mean both \textit{fathers} or \textit{parents}. Some speakers also use the singular form to mean \textit{parent} (and not necessarily \textit{father}). The second limitation applies to languages that are closer to the synthetic end of the analytic-synthetic spectrum; i.e. languages that are agglutinative or highly fusional (e.g., Zulu, Uzbek, Estonian). This approach may not allow for the detection of many agglutinated or fused word forms.  Finally, due to the templated, context-free nature of the lexicon, one term was particularly difficult to disambiguate: \textit{veteran}, which can be used to refer to a soldier or a seasoned professional.\footnote{\url{https://www.merriam-webster.com/dictionary/veteran} retrieved 2023-07-24} Cultural differences also had to be considered in addition to the above ambiguity; for example, Japanese translators mentioned the fact that the Japanese equivalent of the term was infrequently used with the first meaning cited above.\footnote{See \url{  https://en.wikipedia.org/wiki/Article\_9\_of\_the\_Japanese\_Constitution} retrieved 2023-07-24}

\paragraph{Lexicon statistics} In Figure \ref{fig:languages_barplot} we can see the obtained data distribution across number and gender for the different languages. We notice a few outliers. As described above, translators are asked to provide relevant morphologically derived forms. This makes the number of nouns in Estonian to be 7 times larger than the average. For instance, ``woman'' is translated into \textit{naine} ``a woman'', \textit{naise} ``of a woman'', \textit{naisele} ``to a woman'', etc.

\subsection{Proposed Pipeline}
\label{sec:pipeline}

Figure \ref{fig:pipeline_diagram} shows a diagram of the \pipeline pipeline. In the first stage or the counts collection, we work at the sentence level for the NTREX and FLORES-200 and at the document level for Common Crawl.  We segment each sample at the word level using \texttt{Stanza} tokenizer available in the given language \citep{qi-etal-2020-stanza} except for Cantonese (yue) for which we reuse the model available for simplified Chinese (zh-hans) and Thai for which we use \texttt{PyThaiNLP}.\footnote{\url{https://pythainlp.github.io/docs/2.0/api/tokenize.html}} For the rest of the languages we use simple \texttt{nltk}\footnote{\url{https://www.nltk.org/api/nltk.tokenize.html}} typographic tokenizer (based on white-space and punctuation marks). We then count and increment a gender-class counter anytime we match a word in the list of words representative of this class. For instance, in the sentence ``my mother was a nurse'' %
the pipeline will add +1 to the feminine counter (due to lexical match of ``mother''). 

Once this process has been done for each sentence in the dataset we move to the second stage or the reporting of gender proportions where we define a score for each gender-class by dividing the gender-class count by the total number of words in the dataset. By doing so, the final gender score does not depend on any defined linguistic macro-unit such as sentences or documents lengths but only on the word-level tokenization.

\begin{figure}[t!]
    \centering
    \includegraphics[width=0.45\textwidth]{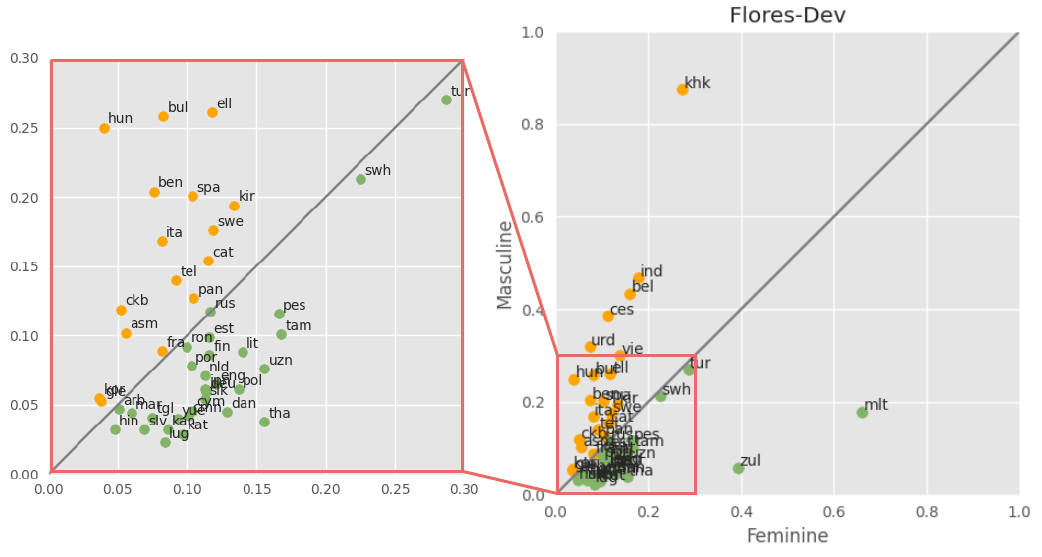} 
    \caption{Gender Representation in \% of the total tokens in the FLORES dataset dev split.}
    \label{fig:devflores_results}
\end{figure}

\begin{figure}[t!]
    \centering
    \includegraphics[width=0.45\textwidth]{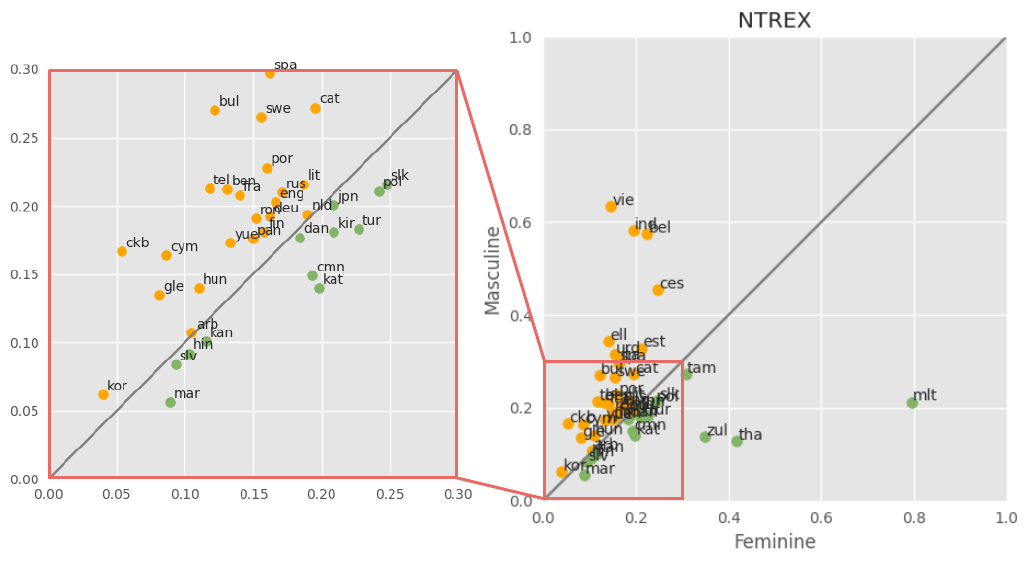} 
    \caption{Gender Representation in \% of the total tokens in the NTREX dataset.}
    \label{fig:NTREX_results}
\end{figure}

\begin{figure}[t!]
    \centering
    \includegraphics[width=0.45\textwidth]{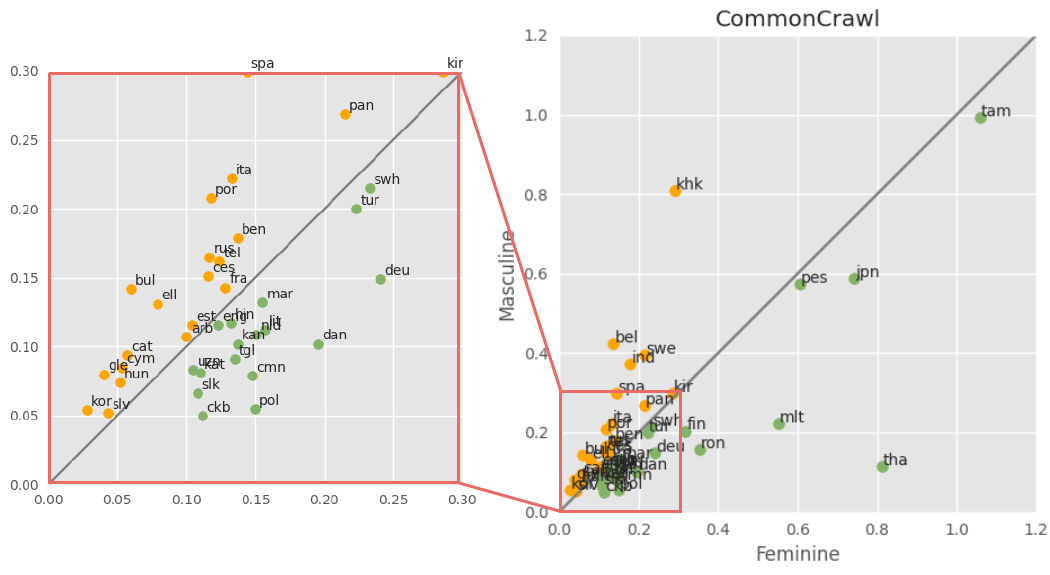} 
    \caption{Gender Representation in \% of the total tokens in the Common Crawl dataset.}
    \label{fig:commoncrawl_results}
\end{figure}

\section{Experiments}
\label{sec:experiments}

To showcase \pipeline, we run it on Common Crawl raw data and two popular machine translation evaluation datasets: FLORES-200 \citep{nllb2022} and NTREX-128 \citep{federmann-etal-2022-ntrex}. FLORES is a Wikipedia-based dataset including 3001 sentences translated from English to 200 languages. NTREX-128 is made of 1997 sentences from news documents originally collected for WMT 2019 \citep{barrault-etal-2019-findings} translated from English into 128 languages. Both these datasets are part of the corpora provided by the WMT shared task. In addition, we run the pipeline on a sample of Common Crawl.\footnote{\url{https://commoncrawl.org/}} Common Crawl is a snapshot of crawlable web data that is widely used in the NLP community thanks to the release of the CCNET corpora \citep{wenzek-etal-2020-ccnet}, the OSCAR corpus \citep{OrtizSuarezSagotRomary2019} and the C4 corpus \citep{2019t5}. It is used to train NLP systems like language and machine translation models. We run it on 100k documents for each language.  
Our pipeline supports \langcount{} languages, and we run it on the intersection of these datasets with the set of supported languages. 

\begin{table}[t!]
\centering\footnotesize
\resizebox{\linewidth}{!}{
\begin{tabular}{c c c c c r} 
 \toprule
 \bf Lang & \textbf{Fem.} & \textbf{Masc.} & \textbf{Uns.} & \bf $\Delta$(|Fem.-Masc.|)  & \textbf{\%doc.} \\ 
 \midrule
 \multicolumn{6}{c}{\textit{Flores DevTest. }}\\
   eng &  \underline{0.121} &   0.065  & \bf 0.379& 0.056 (0.0003)  & 11.2\\

avg. &  0.128 &  \underline{0.144}  &  \bf 0.302 & 0.097 (0.0003)   & 10.1\\

 \midrule
 \multicolumn{6}{c}{\textit{NTREX  }}\\
  eng &  0.166 &   \underline{0.203}  & \bf 0.379& 0.037 (0.0003)  & 15.5\\

avg. &  0.180 &  \underline{0.224}  &  \bf 0.329 & 0.099 (0.0003)   & 13.4\\

\midrule

\multicolumn{6}{c}{\textit{CommonCrawl}}\\
    eng &  \underline{0.120} &   0.115  & \bf 0.243& 0.005 (0.0000)  & 9.4\\
avg. &  0.212 &  \bf 0.260  &  \underline{0.251} & 0.136 (0.0003)   & 12.0\\

\bottomrule
\end{tabular}
}
\caption{\% Gender Distribution in WMT Evaluation dataset. We report the English distribution and the average across all languages (standard deviation indicated between parenthesis). The full table is available in the appendix Table~\ref{table:gender_dist_flores}-\ref{table:gender_dist_cc}. We \textbf{bold} the most represented gender class, and \underline{underline} the second most represented gender class.  We define the the gender gap $\Delta$ defined as the absolute difference between the Feminine and Masculine scores. \%doc. refers to Coverage.
}
\label{table:gender_dist_mean}
\end{table}

\section{Analysis}

\subsection{Quantitative Analysis}
\label{sec:quantitative_analyais}
We report the average coverage and gender distribution in Table~\ref{table:gender_dist_mean} along with the complete tables for the 55 languages in Table~\ref{table:gender_dist_flores}-\ref{table:gender_dist_cc}.

\paragraph{Coverage} We first look at the number of samples for which at least one noun is found (cf. \%doc in Table~\ref{table:gender_dist_mean}). We find that, on average, about 10\% of samples match with at least a noun (between 10.1 and 13.4\% depending on the dataset). We find that the coverage is the largest for Vietnamese (with up 45.7\% of samples matched) and Thai (28.9\% of samples matched) and the smallest for Korean (between 1.7\% and 2.5\% depending on the dataset). This shows that even though our lexicon is restricted to person nouns and kinship relationships, we are still covering a very large number of samples based on which we measure gender representations.

\begin{table*}[t!]
  \centering
  \begin{tabular}{@{}p{15.5cm}@{\hspace{0.2cm}}p{20.5cm}@{}}
   \toprule
     \textbf{Sentence 1: Omission of words/lexical variation}\\
\midrule
    \small Eng: shark injures 13-year-old on lobster dive in california\texttt{~~~~~~~~~~~~~~~~~~~~~~~~~~~~~~~~~~~~~~~~~~~~~~\textbf{masc.+= 0}}\\
   \small Spa: tiburón hiere a un \textbf{niño} de 13 años que buceaba en busca de langostas en california \texttt{~~~~~~~~~~~~~~~~~~~~\textbf{masc.+= 1}}\\ 
   \small Cat: un tauró fereix un \textbf{nen} de 13 anys mentre buscava llagostes a califòrnia \texttt{~~~~~~~~~~~~~~~~~~~~~~~~~~~~~~~~\textbf{masc.+= 1}}\\
    \midrule
    \textbf{Sentence 2: Multiple translations and variation in part of speech}\\
  \midrule
    \small Eng: [...] something increasingly demanded by younger shoppers.\texttt{~~~~~~~~~~~~~~~~~~~~~~~~~~~~~~\textbf{unspecified.+= 0}} \\
   \small Cat: [...]  un aspecte cada cop més demanat pels consumidors més \textbf{joves}. \texttt{~~~~~~~~~~~~~~~~~~~~~~~~~~~\textbf{unspecified.+= 1}} \\ 
   \midrule
   \textbf{Sentence 3: Robust to typographic differences}\\
       \small Eng: \textbf{mother}-of-three willoughby and \textbf{husband} dan baldwin have been close to jones and his \textbf{wife} \texttt{~\textbf{fem.+= 2,masc.+= 1}}\\
       \begin{CJK*}{UTF8}{gbsn} %
\small Cmn: [...]个孩子的\underline{母亲}的威洛比及其\underline{丈夫} dan baldwin 十年来与琼斯及其\underline{妻子} tara保持[...]\texttt{~~~\textbf{fem.+= 2,masc.+= 1}}
\end{CJK*} 
\\
\midrule
\textbf{Sentence 4: Synonyms}\\
\small Eng: [...] the owner of the lloyds pharmacy chain, for £125m, three years ago.\texttt{~~~~~~~~~~~~~~~~~~~~~~~~~~~~~~~~\textbf{masc.+= 0}}\\
\small Vie: [...] ch\`{u} s\~{o} h\~{u}u c\~{u}a chu\~{o}i nh\`{a} thu\`{o}c lloyds, v\'{o}i gi\'{a} 125 tri\~{e}u b\`{a}ng vào \textbf{ba} năm tr\`{u}\'{o}\v{z}c. \texttt{ \textbf{~~~~~~~~~~~~~~~~~~~masc.+= 1}}\\
\bottomrule
    \end{tabular}
    \caption{Selected examples of gender representation across parallel sentences between English and multiple target languages (based on the NTREX dataset). Detected gendered nouns in bold/underlined. We indicate the  counter incremented by the pipeline for the three gender classes (feminine, masculine and unspecified) next to each sentence when there is at least a match in one of the languages. %
}
    \label{fig:examples}
  \end{table*}

\paragraph{Gender Distribution}
Table~\ref{table:gender_dist_mean} shows gender representation for masculine, feminine and unspecified. For better visualization, Figures \ref{fig:devflores_results}, \ref{fig:NTREX_results} and \ref{fig:commoncrawl_results} report the \% of masculine and feminine representation of the total tokens in FLORES, NTREX, and Common Crawl respectively.

On average, the masculine gender is more represented than the feminine in all three datasets. 
Accounting for uncertainty, using the standard deviation to define a confidence interval,\footnote{For a given language, we consider a gender gap between the masculine and feminine genders when the $\Delta$(|Fem.-Masc.|) is higher than two times the standard error. Otherwise, we consider the dataset to be gender balanced.} we find that 30/45 languages are skewed toward the masculine gender for NTREX. This includes languages like English, Arabic, French, Spanish, Vietnamese, and Panjabi. The rest of the languages are either balanced between masculine and feminine (i.e.$\Delta$(|Fem.-Masc.|) is inferior to two times the confidence interval length) or skewed toward the feminine gender.
In addition, we find 16/54 languages skewed toward the masculine gender for all three datasets suggesting an inherent gender bias in these languages. This includes several romance languages such as Spanish, French, Catalan and Italian along with Belarusian, Indonesian, and Panjabi.  %

\paragraph{Impact of Domains}
We find that 14/55 languages for which, the gender representation changes drastically across the different datasets. For instance, the gender differences are much larger in NTREX than in Common Crawl data. More specifically, in Lithuanian the distribution is skewed toward the masculine class for NTREX data, while it is skewed toward the feminine for Common Crawl data. For Danish, the gender representation is balanced for NTREX but skewed toward the Feminine class for Common Crawl data.
This shows that domains highly impact gender representation. NTREX is based on news data, while Common Crawl includes a large diversity of domains from the Web. 
\paragraph{Comparing Genders across Languages}

In addition, we find a large variability across languages. Some languages like Belarus (\texttt{bel}) and Swedish (\texttt{swe}) are highly skewed toward the Masculine gender class, while other languages are much more balanced such as Mandarin Chinese (\texttt{cmn}) or Hindi~(\texttt{hin}). 

We note that gender distribution cannot be compared across languages quantitatively. Indeed, first, our lexicon is based by design on nouns that are not entirely parallel across languages. Second, our metric highly depends on the number of words in each dataset, which is not comparable across all languages due to their differences in morphology and syntax. 
However, as discussed below (\S~\ref{sec:qualitative_analysis}), our pipeline allows us to highlight qualitative differences in how gender is encoded in different languages.

\subsection{Qualitative Analysis: Gender representation variation in parallel data}
\label{sec:qualitative_analysis}
To understand the cause of these gender representation differences across languages, we present several examples in 
Table~\ref{fig:examples}.  %

\begin{itemize}
    \item Omission of words: When comparing English with Romance languages, we observe cases where the gendered word is omitted in English while being translated as a masculine noun in the target language, like Spanish or Catalan. This leads to larger gender representation gaps in these languages.
    \item Multiple translations and part-of-speech: Sentence 2 shows the impact of how a single English word corresponds to multiple words in other languages. The unspecified word "kid" is translated in 10 words in Catalan: unspecified "jove, criatura"; feminine "minyona, menuda, nena, marreca"; masculine, "minyó, menut, nen, marrec", augmenting the coverage in that second language. In addition, some words in Catalan have multiple part-of-speech, like "jove, menuda, menut" which can act as nouns or adjectives.
    \item Sentence 3 illustrates that even with typologically different languages such as English and Mandarin Chinese, our lexical matching approach successfully highlights cases where gender is preserved across languages. 
    \item Finally, in Sentence 4, we illustrate the limit of the context-free approach. Indeed, the noun ``ba'' means both \textit{father} and \textit{three} in Vietnamese, leading to over-estimating the masculine class on some samples. 
 \end{itemize}

In summary, the differences in gender representation across languages point to four distinct phenomena—first, the inherent limit of our context-free lexical approach. Gender is, in some cases, incorrectly estimated by a by-design restricted lexical-matching method (e.g., Sentence 4). Second, different domain distributions may lead to diverse gender representation. As reported in the previous section, for some languages, the gender scores highly vary depending on the domains (e.g., News vs. Web crawled data). This suggests that when we analyze non-parallel data, the domain may be a prevalent factor that explains gender representation differences across languages. Third, as we observe when analyzing parallel data, gender representation differences may come from biases in the translation itself. For instance, in Sentence 1, the translation explicitly encoded the masculine gender in Spanish and Catalan while being gender unspecified in English. Other translations could have preserved the gender. Fourth, the way gender is encoded is, partly at least, unique to each language. Some languages are inherently biased toward the masculine gender (e.g. ``padres'', which may mean both \textit{fathers} and \textit{parents} in Spanish). Other languages do not always have genderless nouns. For instance, \textit{siblings} can only be translated onto Lithuanian as ``broliai ir seserys'' \textit{Brothers and Sisters}.

\section{Conclusion}

In this work, we presented \pipeline, a large scale multilingual pipeline to compute gender distribution across 55 languages. We find that broadly used datasets are biased toward masculine gender.  Based on this finding, our primary recommendation for multilingual NLP practitioner is to report the gender distribution along with the performance score. This allows reader and systems adopters to be aware of these biases in order to integrate this in their system deployment. Secondly, based on our multilingual lexicon, many directions could be taken to mitigate biases in the performance of the systems (due to biases in the data). \citet{qian-etal-2022-perturbation} developed a perturbation-based technique to build NLP systems that are less biased toward specific group. We envision using our multilingual lexicon to adapt this technique beyond English.

\section*{Limitations and Ethical Statement}

\textbf{English-centric} We designed the list of gendered nouns starting from the English language and then scaled it to multiple languages. This means that our approach may cover incompletely the nuances in different language families regarding gender or only cover them partially and from an English-centric perspective.

\paragraph{Non-Binary Gender Modeling}
To favor scalability across 55 languages, we chose to use a three gender class lexicon. However, this  restrict our approach to binary genders (masculine and feminine) and we only measure imperfectly non-binary genders distribution \cite{haynes2001unseen,herdt2020third} with the ``unspecified'' class. We leave for future work the refinement of our lexical categories in order to measure more granularly genders across languages. 

\paragraph{Lexical Matching} The core assumption of this work is that our predefined lexicon defined in Section~\ref{sec:lexicon} gives us a proxy to account for gender distributions in large datasets. Although our lexicon is obviously not exhaustive, it is simple enough to scale to highly multilingual environments. Future work could consider other types of nouns (beyond family relations or persons) such as gendered occupations nouns, pronouns, etc.

\section*{Acknowledgements}

We thank Mark Tygert for his help and feedback on the statistical analysis, Carleigh Wood for her help with the translation and Gabriel Mejia Gonzalez for his help with linguistic analysis.

\bibliography{anthology,custom}
\bibliographystyle{acl_natbib}

\appendix
\begin{table*}[h!]
\centering\small
\begin{tabular}{c c c c c c c} 
 \toprule
 \bf Lang & \textbf{Feminine} & \textbf{Masculine} & \textbf{Unspecified}   & \bf $\Delta$(|~Fem.-Masc.~|) (ste.) & \textbf{\# words}& \textbf{\% matched sentences} \\ 
 \midrule
 \multicolumn{7}{c}{\textit{Flores DevTest.}}\\
    eng &  \underline{0.121} &   0.065  & \bf 0.379& 0.056 (0.0003) & 23211 & 11.2\\

  arb &  \underline{0.051} &   0.047  & \bf 0.094& 0.004 (0.0002) & 25549 & 4.1\\
  asm &  0.056 &   \bf 0.102  & \underline{0.093}& 0.046 (0.0003) & 21610 & 4.5\\
  bel &  0.161 &   \underline{0.434}  & \bf 0.444& 0.274 (0.0005) & 21174 & 12.7\\
  ben &  0.076 &   \bf 0.204  & \underline{0.142}& 0.128 (0.0004) & 21101 & 7.2\\
  bul &  0.083 &   \bf 0.258  & \underline{0.114}& 0.175 (0.0004) & 22834 & 9.1\\
  cat &  0.115 &   \bf 0.154  & \underline{0.146}& 0.038 (0.0003) & 26005 & 9.4\\
  ces &  0.113 &   \bf 0.385  & \underline{0.153}& 0.271 (0.0005) & 20284 & 10.6\\
  ckb &  0.052 &   \underline{0.119}  & \bf 0.152& 0.066 (0.0003) & 21073 & 4.3\\
  cmn &  \underline{0.101} &   0.042  & \bf 0.794& 0.059 (0.0002) & 23676 & 17.6\\
  cym &  \underline{0.104} &   0.046  & \bf 0.146& 0.058 (0.0002) & 26013 & 6.4\\
  dan &  \underline{0.129} &   0.045  & \bf 0.160& 0.085 (0.0003) & 22471 & 6.3\\
  deu &  \underline{0.114} &   0.059  & \bf 0.301& 0.055 (0.0003) & 21922 & 9.2\\
  ell &  0.118 &   \bf 0.261  & \underline{0.253}& 0.143 (0.0004) & 24548 & 12.8\\
  est &  \underline{0.116} &   0.099  & \bf 0.519& 0.017 (0.0003) & 18107 & 11.0\\
  fin &  \underline{0.116} &   0.086  & \bf 0.147& 0.031 (0.0004) & 16314 & 4.9\\
  fra &  0.082 &   \underline{0.089}  & \bf 0.234& 0.007 (0.0003) & 26910 & 9.6\\
  gle &  0.038 &   \underline{0.053}  & \bf 0.479& 0.015 (0.0002) & 26517 & 12.3\\
  hin &  \underline{0.048} &   0.032  & \bf 0.104& 0.016 (0.0002) & 25094 & 3.8\\
  hun &  0.040 &   \bf 0.250  & \underline{0.060}& 0.210 (0.0004) & 19977 & 6.0\\
  ind &  0.179 &   \bf 0.468  & \underline{0.193}& 0.289 (0.0006) & 20728 & 14.5\\
  ita &  0.082 &   \underline{0.168}  & \bf 0.223& 0.086 (0.0003) & 25583 & 10.2\\
  jpn &  \underline{0.113} &   0.061  & \bf 0.716& 0.052 (0.0002) & 31000 & 20.4\\
  kan &  \underline{0.086} &   0.032  & \bf 0.102& 0.054 (0.0002) & 18593 & 3.1\\
  kat &  \bf 0.097 &   0.029  & \underline{0.068}& 0.068 (0.0002) & 20527 & 3.0\\
  khk &  \underline{0.274} &   \bf 0.874  & 0.270& 0.599 (0.0007) & 21861 & 22.6\\
  kir &  0.134 &   \underline{0.194}  & \bf 0.482& 0.060 (0.0004) & 20120 & 12.7\\
  kor &  \underline{0.037} &   \bf 0.055  & 0.012& 0.018 (0.0002) & 16341 & 1.7\\
  lit &  \bf 0.140 &   0.088  & \underline{0.125}& 0.052 (0.0003) & 19246 & 5.4\\
  lug &  \underline{0.084} &   0.023  & \bf 0.606& 0.061 (0.0002) & 21457 & 12.6\\
  mar &  \bf 0.060 &   0.044  & \underline{0.055}& 0.016 (0.0002) & 18281 & 2.5\\
  mlt &  \bf 0.661 &   0.179  & \underline{0.191}& 0.482 (0.0005) & 25104 & 18.3\\
  nld &  \underline{0.113} &   0.071  & \bf 0.236& 0.042 (0.0003) & 21229 & 7.5\\
  pan &  \underline{0.105} &   \bf 0.127  & 0.087& 0.022 (0.0003) & 27651 & 6.5\\
  pes &  \underline{0.166} &   0.116  & \bf 0.310& 0.050 (0.0003) & 24157 & 10.0\\
  pol &  \underline{0.137} &   0.061  & \bf 0.544& 0.076 (0.0003) & 21143 & 13.4\\
  por &  \underline{0.103} &   0.078  & \bf 0.338& 0.025 (0.0003) & 24269 & 10.9\\
  ron &  \underline{0.100} &   0.092  & \bf 0.240& 0.008 (0.0003) & 25046 & 8.9\\
  rus &  \underline{0.117} &   \bf 0.117  & 0.098& 0.000 (0.0003) & 21431 & 5.4\\
  slk &  \underline{0.113} &   0.054  & \bf 0.508& 0.059 (0.0003) & 20292 & 11.5\\
  slv &  \underline{0.069} &   0.032  & \bf 0.069& 0.037 (0.0002) & 21586 & 3.3\\
  spa &  0.104 &   \underline{0.201}  & \bf 0.260& 0.097 (0.0003) & 26896 & 12.3\\
  swe &  0.119 &   \underline{0.176}  & \bf 0.200& 0.057 (0.0004) & 20969 & 8.9\\
  swh &  \underline{0.225} &   0.213  & \bf 0.689& 0.013 (0.0004) & 23964 & 20.4\\
  tam &  \bf 0.168 &   0.101  & \underline{0.123}& 0.067 (0.0003) & 17862 & 4.5\\
  tel &  0.092 &   \bf 0.140  & \underline{0.122}& 0.049 (0.0004) & 16373 & 3.9\\
  tgl &  \underline{0.075} &   0.041  & \bf 0.373& 0.034 (0.0002) & 29518 & 11.1\\
  tha &  \underline{0.156} &   0.038  & \bf 0.439& 0.118 (0.0003) & 28922 & 12.7\\
  tur &  \underline{0.287} &   0.270  & \bf 0.293& 0.017 (0.0005) & 17775 & 8.4\\
  urd &  0.074 &   \bf 0.320  & \underline{0.234}& 0.245 (0.0004) & 26887 & 9.2\\
  uzn &  \underline{0.156} &   0.076  & \bf 0.260& 0.080 (0.0003) & 21181 & 8.3\\
  vie &  0.139 &   \underline{0.301}  & \bf 1.441& 0.162 (0.0004) & 25263 & 30.6\\
  yue &  \underline{0.093} &   0.040  & \bf 0.837& 0.053 (0.0002) & 24728 & 19.1\\
  zul &  \underline{0.394} &   0.059  & \bf 0.653& 0.335 (0.0005) & 18532 & 17.0\\
 \cdashline{1-7}
avg. &  0.128 &  \underline{0.144}  &  \bf 0.302 & 0.097 (0.0003)  &  22572 & 10.1\\

\bottomrule
\end{tabular}
\caption{\% Gender Distribution in FLORES-200  dataset \citep{nllb2022}. We \textbf{bold} the most represented gender class, and \underline{underline} the second most represented gender class for each language. We report $\Delta$ the gender gap defined as the absolute difference between the Feminine and Masculine scores along with the standard error (ste.). \% matched sentences refers to the coverage of our pipeline (cf. \S~\ref{sec:quantitative_analyais}). %
}
\label{table:gender_dist_flores}
\end{table*}
\begin{table*}[h!]
\centering\small
\begin{tabular}{c c c c c c c} 
 \toprule
 \bf Lang & \textbf{Feminine} & \textbf{Masculine} & \textbf{Unspecified}   & \bf $\Delta$(|~Fem.-Masc.~|) (ste.) & \textbf{\# words}& \textbf{\% matched sentences} \\ 
 \midrule
 \multicolumn{7}{c}{\textit{NTREX}}\\
  eng &  0.166 &   \underline{0.203}  & \bf 0.379& 0.037 (0.0003) & 48254 & 15.5\\

  arb &  0.105 &   \underline{0.107}  & \bf 0.206& 0.002 (0.0002) & 51388 & 8.7\\
  bel &  0.224 &   \bf 0.574  & \underline{0.397}& 0.350 (0.0004) & 44597 & 16.9\\
  ben &  0.131 &   \underline{0.212}  & \bf 0.311& 0.081 (0.0003) & 40505 & 11.3\\
  bul &  \underline{0.122} &   \bf 0.270  & 0.095& 0.148 (0.0003) & 49283 & 10.5\\
  cat &  0.195 &   \bf 0.272  & \underline{0.235}& 0.077 (0.0003) & 54401 & 15.6\\
  ces &  \underline{0.248} &   \bf 0.454  & 0.190& 0.206 (0.0004) & 43623 & 16.3\\
  ckb &  0.054 &   \underline{0.167}  & \bf 0.244& 0.113 (0.0002) & 42554 & 6.5\\
  cmn &  \underline{0.193} &   0.149  & \bf 0.944& 0.044 (0.0003) & 50326 & 24.8\\
  cym &  0.086 &   \bf 0.164  & \underline{0.154}& 0.078 (0.0002) & 52540 & 8.8\\
  dan &  \underline{0.184} &   0.177  & \bf 0.186& 0.007 (0.0003) & 45684 & 10.7\\
  deu &  0.162 &   \underline{0.192}  & \bf 0.276& 0.030 (0.0003) & 46398 & 12.3\\
  ell &  0.141 &   \bf 0.344  & \underline{0.170}& 0.203 (0.0003) & 51204 & 14.4\\
  est &  0.212 &   \underline{0.328}  & \bf 0.458& 0.116 (0.0004) & 37794 & 15.8\\
  fin &  0.158 &   \underline{0.181}  & \bf 0.196& 0.024 (0.0003) & 33617 & 7.9\\
  fra &  0.140 &   \underline{0.208}  & \bf 0.258& 0.068 (0.0003) & 54336 & 13.9\\
  gle &  0.081 &   \underline{0.135}  & \bf 0.493& 0.054 (0.0002) & 54205 & 16.2\\
  hin &  \underline{0.103} &   0.092  & \bf 0.147& 0.011 (0.0002) & 55207 & 8.1\\
  hun &  \underline{0.110} &   \bf 0.140  & 0.072& 0.030 (0.0002) & 42834 & 6.6\\
  ind &  0.195 &   \bf 0.581  & \underline{0.213}& 0.386 (0.0004) & 45071 & 18.1\\
  ita &  0.166 &   \bf 0.301  & \underline{0.229}& 0.135 (0.0003) & 51884 & 14.8\\
  jpn &  \underline{0.209} &   0.201  & \bf 0.868& 0.008 (0.0002) & 59704 & 25.2\\
  kan &  \underline{0.115} &   0.101  & \bf 0.131& 0.014 (0.0002) & 36574 & 4.9\\
  kat &  \bf 0.198 &   \underline{0.140}  & 0.103& 0.058 (0.0002) & 39912 & 5.1\\
  kir &  \underline{0.209} &   0.181  & \bf 0.310& 0.028 (0.0003) & 38682 & 12.0\\
  kor &  0.040 &   \bf 0.062  & \underline{0.059}& 0.022 (0.0002) & 32204 & 2.5\\
  lit &  \underline{0.187} &   \bf 0.216  & 0.153& 0.029 (0.0003) & 41190 & 9.4\\
  mar &  \bf 0.089 &   0.056  & \underline{0.069}& 0.033 (0.0002) & 35980 & 3.6\\
  mlt &  \bf 0.795 &   0.212  & \underline{0.284}& 0.583 (0.0004) & 51466 & 24.7\\
  nld &  0.190 &   \underline{0.194}  & \bf 0.196& 0.004 (0.0003) & 48003 & 11.2\\
  pan &  \underline{0.150} &   \bf 0.176  & 0.100& 0.026 (0.0002) & 53845 & 9.9\\
  pol &  \underline{0.242} &   0.211  & \bf 0.525& 0.030 (0.0003) & 42638 & 17.9\\
  por &  0.160 &   \underline{0.228}  & \bf 0.244& 0.067 (0.0003) & 50482 & 13.8\\
  ron &  0.152 &   \underline{0.191}  & \bf 0.367& 0.039 (0.0002) & 54463 & 15.5\\
  rus &  \underline{0.171} &   \bf 0.210  & 0.089& 0.039 (0.0003) & 46295 & 8.5\\
  slk &  \underline{0.248} &   0.216  & \bf 0.420& 0.033 (0.0003) & 43063 & 16.0\\
  slv &  \bf 0.093 &   \underline{0.084}  & 0.077& 0.009 (0.0002) & 45339 & 4.8\\
  spa &  0.162 &   \underline{0.297}  & \bf 0.344& 0.135 (0.0003) & 52579 & 15.9\\
  swe &  0.156 &   \bf 0.265  & \underline{0.240}& 0.109 (0.0003) & 42980 & 12.3\\
  tam &  \bf 0.308 &   \underline{0.273}  & 0.068& 0.035 (0.0002) & 36960 & 7.0\\
  tel &  \underline{0.118} &   \bf 0.213  & 0.086& 0.095 (0.0003) & 31427 & 5.0\\
  tha &  \underline{0.418} &   0.128  & \bf 0.870& 0.290 (0.0003) & 57923 & 23.1\\
  tur &  \underline{0.227} &   0.183  & \bf 0.252& 0.044 (0.0003) & 36163 & 8.1\\
  vie &  0.146 &   \underline{0.633}  & \bf 2.166& 0.487 (0.0004) & 52577 & 45.7\\
  yue &  0.133 &   \underline{0.173}  & \bf 0.933& 0.041 (0.0002) & 54233 & 26.6\\
   \cdashline{1-7}

avg. &  0.180 &  0.224  &  0.329 & 0.099 (0.0003)  &  46231 & 13.4\\

\bottomrule
\end{tabular}
\caption{\% Gender Distribution in NTREX data \citep{federmann-etal-2022-ntrex}. We \textbf{bold} the most represented gender class, and \underline{underline} the second most represented gender class for each language. We report $\Delta$ the gender gap defined as the absolute difference between the Feminine and Masculine scores along with the standard error (ste.). \% matched sentences refers to the coverage of our pipeline (cf. \S~\ref{sec:quantitative_analyais}). %
}
\label{table:gender_dist_ntrex}
\end{table*}
\begin{table*}[h!]
\centering\small
\begin{tabular}{c c c c c c c} 
 \toprule
 \bf Lang & \textbf{Feminine} & \textbf{Masculine} & \textbf{Unspecified}   & \bf $\Delta$(|~Fem.-Masc.~|) (ste.) & \textbf{\# words}& \textbf{\% matched documents} \\ 
 
 \midrule
 \multicolumn{7}{c}{\textit{CommonCrawl}}\\
     eng &  \underline{0.120} &   0.115  & \bf 0.243& 0.005 (0.0000) & 2529756 & 9.4\\

  arb &  \underline{0.101} &   \bf 0.106  & 0.085& 0.005 (0.0000) & 6078083 & 9.5\\
  bel &  0.122 &   \bf 0.447  & \underline{0.358}& 0.325 (0.0000) & 2430561 & 14.1\\
  ben &  \underline{0.158} &   \bf 0.199  & 0.140& 0.041 (0.0000) & 4603054 & 14.4\\
  bul &  0.072 &   \bf 0.145  & \underline{0.142}& 0.073 (0.0000) & 2708232 & 7.7\\
  cat &  0.079 &   \underline{0.141}  & \bf 0.152& 0.062 (0.0000) & 3157729 & 9.1\\
  ces &  0.117 &   \underline{0.146}  & \bf 0.165& 0.030 (0.0000) & 2366804 & 7.9\\
  ckb &  \underline{0.108} &   0.049  & \bf 0.124& 0.059 (0.0000) & 5341945 & 10.2\\
  cmn &  \underline{0.170} &   0.097  & \bf 0.519& 0.072 (0.0000) & 5484451 & 23.8\\
  cym &  0.079 &   \underline{0.082}  & \bf 0.164& 0.003 (0.0000) & 2777579 & 7.4\\
  dan &  \underline{0.182} &   0.102  & \bf 0.201& 0.080 (0.0000) & 2310993 & 7.9\\
  deu &  \underline{0.144} &   0.099  & \bf 0.187& 0.044 (0.0000) & 2148705 & 6.8\\
  ell &  0.068 &   \bf 0.143  & \underline{0.142}& 0.075 (0.0000) & 2855903 & 7.7\\
  est &  0.112 &   \underline{0.152}  & \bf 0.429& 0.040 (0.0000) & 1943773 & 10.3\\
  fin &  \bf 0.294 &   \underline{0.201}  & 0.155& 0.094 (0.0001) & 1621020 & 7.3\\
  fra &  0.110 &   \underline{0.136}  & \bf 0.151& 0.025 (0.0000) & 2857434 & 8.3\\
  gle &  0.044 &   \underline{0.101}  & \bf 0.406& 0.057 (0.0000) & 2634719 & 12.2\\
  hin &  \bf 0.176 &   \underline{0.124}  & 0.065& 0.052 (0.0000) & 2675603 & 7.4\\
  hun &  0.058 &   \bf 0.097  & \underline{0.075}& 0.038 (0.0000) & 2572506 & 4.5\\
  ind &  0.183 &   \bf 0.367  & \underline{0.184}& 0.185 (0.0000) & 2227691 & 12.1\\
  ita &  \underline{0.131} &   \bf 0.195  & 0.070& 0.064 (0.0000) & 2961219 & 8.2\\
  jpn &  \underline{0.858} &   0.724  & \bf 0.963& 0.134 (0.0000) & 5964414 & 27.4\\
  kan &  \bf 0.103 &   \underline{0.094}  & 0.093& 0.009 (0.0000) & 3772755 & 7.0\\
  kat &  \bf 0.129 &   0.089  & \underline{0.116}& 0.040 (0.0000) & 3977699 & 6.2\\
  khk &  \underline{0.301} &   \bf 0.948  & 0.248& 0.647 (0.0000) & 4996882 & 32.1\\
  kir &  0.269 &   \bf 0.308  & \underline{0.270}& 0.039 (0.0000) & 3895597 & 20.0\\
  kor &  0.032 &   \underline{0.047}  & \bf 0.047& 0.015 (0.0000) & 2364450 & 2.4\\
  lit &  \underline{0.148} &   0.117  & \bf 0.243& 0.031 (0.0000) & 2293338 & 8.4\\
  mar &  \bf 0.133 &   \underline{0.112}  & 0.051& 0.021 (0.0000) & 1531197 & 3.8\\
  mlt &  \bf 0.554 &   0.179  & \underline{0.213}& 0.375 (0.0001) & 2437212 & 20.0\\
  nld &  \underline{0.127} &   0.101  & \bf 0.201& 0.027 (0.0000) & 1921934 & 6.3\\
  pan &  \underline{0.236} &   \bf 0.308  & 0.074& 0.072 (0.0000) & 6772503 & 22.4\\
  pes &  \underline{1.459} &   1.425  & \bf 1.514& 0.034 (0.0000) & 3881584 & 14.7\\
  pol &  \underline{0.175} &   0.074  & \bf 0.290& 0.101 (0.0000) & 2453053 & 9.9\\
  por &  0.110 &   \bf 0.160  & \underline{0.158}& 0.050 (0.0000) & 2846706 & 9.2\\
  ron &  \underline{0.207} &   0.138  & \bf 0.257& 0.068 (0.0000) & 2555624 & 10.1\\
  rus &  0.107 &   \bf 0.139  & \underline{0.117}& 0.031 (0.0000) & 2565203 & 6.4\\
  slk &  \underline{0.111} &   0.066  & \bf 0.324& 0.045 (0.0000) & 2269033 & 8.7\\
  slv &  0.057 &   \underline{0.071}  & \bf 0.142& 0.014 (0.0000) & 2373967 & 5.3\\
  spa &  0.122 &   \bf 0.255  & \underline{0.183}& 0.133 (0.0000) & 3046193 & 11.7\\
  swe &  \underline{0.179} &   \bf 0.372  & 0.157& 0.193 (0.0000) & 2346273 & 11.5\\
  swh &  \underline{0.221} &   0.194  & \bf 0.492& 0.027 (0.0000) & 2385794 & 19.9\\
  tam &  \bf 0.766 &   \underline{0.676}  & 0.073& 0.091 (0.0000) & 1691612 & 11.7\\
  tel &  \underline{0.127} &   \bf 0.165  & 0.056& 0.038 (0.0000) & 1277513 & 3.5\\
  tgl &  \underline{0.145} &   0.108  & \bf 0.419& 0.038 (0.0000) & 5035687 & 21.2\\
  tha &  \underline{0.735} &   0.107  & \bf 0.932& 0.628 (0.0000) & 7142646 & 28.9\\
  tur &  \bf 0.228 &   0.202  & \underline{0.215}& 0.027 (0.0000) & 2293026 & 7.3\\
  uzn &  \underline{0.119} &   0.077  & \bf 0.280& 0.042 (0.0000) & 2973725 & 9.5\\
  \cdashline{1-7}

avg. &  0.212 &  0.260  &  0.251 & 0.136 (0.0003)  &  3088848 & 12.0\\
\bottomrule
\end{tabular}
\caption{\% Gender Distribution in a Common Crawl sample. We \textbf{bold} the most represented gender class, and \underline{underline} the second most represented gender class. We report $\Delta$ the gender gap defined as the absolute difference between the Feminine and Masculine scores along with the standard error (ste.). \% matched documents refers to the coverage of our pipeline (cf. \S~\ref{sec:quantitative_analyais}).}
\label{table:gender_dist_cc}
\end{table*}

\begin{table*}[h!]
\centering
\small
\begin{tabular}{ll}
\toprule
\bf Language Code & \bf Language\\
\midrule
arb$\_$Arab&	Modern Standard Arabic\\
asm$\_$Beng&	Assamese\\
bel$\_$Cyrl&	Belarusian\\
ben$\_$Beng&	Bengali\\
bul$\_$Cyrl&	Bulgarian\\
cat$\_$Latn&	Catalan\\
ces$\_$Latn&	Czech\\
ckb$\_$Arab&	Central Kurdish\\
cmn$\_$Hans&	Mandarin Chinese (simplified script)\\
cym$\_$Latn&	Welsh\\
dan$\_$Latn&	Danish\\
deu$\_$Latn&	German\\
ell$\_$Grek&	Greek\\
eng$\_$Latn&	English\\
est$\_$Latn&	Estonian\\
fin$\_$Latn&	Finnish\\
fra$\_$Latn&	French\\
gle$\_$Latn&	Irish\\
hin$\_$Deva&	Hindi\\
hun$\_$Latn&	Hungarian\\
ind$\_$Latn&	Indonesian\\
ita$\_$Latn&	Italian\\
jpn$\_$Jpan&	Japanese\\
kat$\_$Geor&	Georgian\\
khk$\_$Cyrl&	Halh Mongolian\\
kir$\_$Cyrl&	Kyrgyz\\
lit$\_$Latn&	Lithuanian\\
lug$\_$Latn&	Ganda\\
lvs$\_$Latn&	Standard Latvian\\
mar$\_$Deva&	Marathi\\
mlt$\_$Latn&	Maltese\\
nld$\_$Latn&	Dutch\\
pan$\_$Guru&	Eastern Panjabi\\
pes$\_$Arab&	Western Persian\\
pol$\_$Latn&	Polish\\
por$\_$Latn&	Portuguese\\
ron$\_$Latn&	Romanian\\
rus$\_$Cyrl&	Russian\\
slk$\_$Latn&	Slovak\\
slv$\_$Latn&	Slovenian\\
spa$\_$Latn&	Spanish\\
swe$\_$Latn&	Swedish\\
swh$\_$Latn&	Swahili\\
tam$\_$Taml&	Tamil\\
tha$\_$Thai&	Thai\\
tur$\_$Latn&	Turkish\\
ukr$\_$Cyrl&	Ukrainian\\
urd$\_$Arab&	Urdu\\
uzn$\_$Latn&	Northern Uzbek\\
vie$\_$Latn&	Vietnamese\\
yue$\_$Hant&	Yue Chinese (traditional script)\\
kan$\_$Knda	& Kannada \\
tel$\_$Telu	& Telugu \\
tgl$\_$Latn	& Tagalog \\
zul$\_$Latn	& Zulu \\
\bottomrule
\end{tabular}
\caption{The 55 languages analyzed in this work, subselected from the 200 \nllb{} languages \citep{nllb2022}. %
}
\label{table:language_list}
\end{table*}

\end{document}